\documentclass[conference]{IEEEtran}
\usepackage{bm}
\IEEEoverridecommandlockouts
\usepackage{cite}
\usepackage{amsmath,amssymb,amsfonts}
\usepackage{algorithmic}
\usepackage{graphicx}
\usepackage{textcomp}
\usepackage{xcolor}
\usepackage{array} 
\usepackage{makecell}
\usepackage{multirow}
\def\BibTeX{{\rm B\kern-.05em{\sc i\kern-.025em b}\kern-.08em
    T\kern-.1667em\lower.7ex\hbox{E}\kern-.125emX}}
\begin{document}

\title{ConfidentSplat: Confidence-Weighted Depth Fusion for Accurate 3D Gaussian Splatting SLAM
{\footnotesize \textsuperscript{}}

\thanks{*Corresponding Author}
}

\author{\IEEEauthorblockN{Amanuel T. Dufera}
\IEEEauthorblockA{\textit{Control Science and Engineering } \\
\textit{Xi'an Jiaotong University }\\
Xi'an, PR of China \\
amani@stu.xjtu.edu.cn}
\and
\IEEEauthorblockN{Yuan-Li Cai \textsuperscript{*}}
\IEEEauthorblockA{\textit{School of Automation Science and Engineering} \\
\textit{Xi'an Jiaotong University}\\
Xi'an, PR of China \\
ylicai@mail.xjtu.edu.cn}
\and

}

\maketitle
\begin{abstract}
We introduce \textit{ConfidentSplat}, a novel 3D Gaussian Splatting (3DGS)-based SLAM system for robust, high-fidelity RGB-only reconstruction. Addressing geometric inaccuracies in existing RGB-only 3DGS SLAM methods that stem from unreliable depth estimation, ConfidentSplat incorporates a core innovation: a confidence-weighted fusion mechanism. This mechanism adaptively integrates depth cues from multi-view geometry with learned monocular priors (Omnidata ViT), dynamically weighting their contributions based on explicit reliability estimates—derived predominantly from multi-view geometric consistency—to generate high-fidelity proxy depth for map supervision. The resulting proxy depth guides the optimization of a deformable 3DGS map, which efficiently adapts online to maintain global consistency following pose updates from a DROID-SLAM-inspired frontend and backend optimizations (loop closure, global bundle adjustment). Extensive validation on standard benchmarks (TUM-RGBD, ScanNet) and diverse custom mobile datasets demonstrates significant improvements in reconstruction accuracy (L1 depth error) and novel view synthesis fidelity (PSNR, SSIM, LPIPS) over baselines, particularly in challenging conditions. ConfidentSplat underscores the efficacy of principled, confidence-aware sensor fusion for advancing state-of-the-art dense visual SLAM.
\end{abstract}


\section{Introduction}

Real-time, high-fidelity 3D scene reconstruction via Simultaneous Localization and Mapping (SLAM) is pivotal for autonomous systems and immersive computing \cite{rs15041156, Lajoie_2022}. The advent of 3D Gaussian Splatting (3DGS) \cite{kerbl3Dgaussians} has revolutionized this domain, offering an explicit, efficient, and highly expressive representation conducive to photorealistic rendering and continuous optimization \cite{huang2024photoslamrealtimesimultaneouslocalization, yugay2024gaussianslamphotorealisticdenseslam, yan2024gsslamdensevisualslam, Matsuki:Murai:etal:CVPR2024}. Consequently, state-of-the-art SLAM systems increasingly leverage 3DGS, often employing robust visual odometry frontends like DROID-SLAM \cite{teed2022droidslamdeepvisualslam} to achieve impressive real-time performance using only RGB input \cite{sandström2024splatslamgloballyoptimizedrgbonly, homeyer2024droidsplatcombiningendtoendslam}.

Despite these strides, ensuring high geometric fidelity within purely RGB-based 3DGS SLAM pipelines remains a critical challenge. The accuracy of reconstructed geometry fundamentally depends on reliable depth supervision. However, depth derived solely from multi-view stereo constraints can be 
\vspace{1em}  
sparse or unreliable in ill-conditioned regions (e.g., texture-poor surfaces), while depth inferred from monocular priors often lacks metric accuracy and global consistency. Existing systems frequently resort to heuristic or overly simplistic strategies for combining these distinct depth sources, failing to rigorously account for their varying reliability across different scene parts or viewpoints. This oversight can lead to geometric inaccuracies and suboptimal reconstruction quality.

To address this fundamental limitation, we introduce \textit{ConfidentSplat}, a novel RGB-only SLAM framework building upon DROID-SLAM \cite{teed2022droidslamdeepvisualslam} and 3DGS \cite{kerbl3Dgaussians}. Our central innovation is a principled \textit{confidence-weighted fusion mechanism} that generates high-fidelity proxy depth maps for supervising the mapping process. This mechanism intelligently integrates dense depth estimates derived from multi-view geometric consistency with scale-aligned predictions from learned monocular priors. Crucially, the fusion process dynamically modulates the contribution of each source based on explicit, quantifiable estimates of their respective reliability at a pixel level. By prioritizing more dependable depth information, ConfidentSplat achieves significantly enhanced geometric accuracy and reconstruction robustness. Furthermore, a deformable 3DGS map representation allows the scene to efficiently adapt online, maintaining global consistency following trajectory corrections from backend optimizations like loop closure and global bundle adjustment.

Our primary contribution is a principled methodology for adaptive proxy depth generation within a 3DGS SLAM pipeline. This employs a novel confidence-weighted fusion strategy to synergistically combine depth from multi-view geometry with estimates from learned monocular priors, dynamically weighting each source by its estimated reliability. We introduce tailored confidence estimation metrics: multi-view depth reliability is assessed via pairwise geometric consistency checks between keyframes (extending \cite{sandström2024splatslamgloballyoptimizedrgbonly}), while monocular prior stability is evaluated using temporal consistency. The resulting higher-fidelity local depth enhances frontend tracking and provides superior geometric supervision for the deformable 3DGS map, critically enabling global consistency after backend optimizations. Comprehensive empirical validation demonstrates substantial quantitative improvements in geometric accuracy (depth error) and novel view synthesis quality (PSNR, SSIM, LPIPS) across standard benchmarks and challenging custom datasets compared to relevant baselines.

\section{Related Work: Dense Visual SLAM and Depth Estimation}

\subsection{Evolution of Dense Visual SLAM}
Dense visual SLAM initially relied on volumetric representations such as Truncated Signed Distance Functions (TSDFs) \cite{10.1145/237170.237269}, popularized by systems like KinectFusion \cite{6162880}. Subsequent advancements focused on scalability and efficiency through techniques like voxel hashing \cite{10.1145/2508363.2508374, 7165673, Oleynikova_2017, matsuki2023newtonneuralviewcentricmapping}, octree-based partitioning \cite{6751517, Yang_2022, Marniok2017AnEO, liu2021neuralsparsevoxelfields}, and point-based fusion \cite{10.1145/3182157, 9591241, 8954208}. Addressing pose drift via global optimization, including submap strategies and loop closure, became critical for large-scale consistency \cite{7299077, dai2017bundlefusionrealtimegloballyconsistent, Khler2016RealTimeLD, maier2017efficientonlinesurfacecorrection, mao2024ngelslamneuralimplicitrepresentationbased, tang2023mipsfusionmultiimplicitsubmapsscalablerobust, liso2024loopyslamdenseneuralslam}. Direct methods, notably DROID-SLAM \cite{teed2022droidslamdeepvisualslam}, significantly refined tracking accuracy by iteratively optimizing optical flow and camera poses within a factor graph, paving the way for highly accurate systems such as GO-SLAM \cite{zhang2023goslamglobaloptimizationconsistent}, HI-SLAM \cite{zhang2023hislammonocularrealtimedense}, and GlORIE-SLAM \cite{zhang2024glorieslamgloballyoptimizedrgbonly}.

\subsection{Neural Implicit Representations and Depth in RGB-Only SLAM}
The advent of Neural Radiance Fields (NeRFs) \cite{mildenhall2020nerfrepresentingscenesneural} catalyzed numerous dense SLAM systems leveraging neural implicit representations from RGB input \cite{rosinol2022nerfslamrealtimedensemonocular, hua2023fmappingfactorizedefficientneural}. Early NeRF-based SLAM methods often faced challenges with global map consistency and computational overhead for map updates \cite{rosinol2022nerfslamrealtimedensemonocular}. While some approaches like NeRF-SLAM \cite{rosinol2022nerfslamrealtimedensemonocular} integrated depth priors (e.g., from single-view estimators or DROID-SLAM \cite{teed2022droidslamdeepvisualslam}), they generally lacked sophisticated mechanisms for fusing multi-view geometric information or explicitly reasoning about depth uncertainty. Systems such as FMapping \cite{hua2023fmappingfactorizedefficientneural} prioritized representational efficiency but primarily relied on photometric consistency, which can be fragile. Even with the emergence of globally consistent methods like GO-SLAM \cite{zhang2023goslamglobaloptimizationconsistent}, robustly estimating and fusing depth from potentially conflicting cues with explicit uncertainty quantification remained a persistent challenge.

\subsection{3DGS in SLAM}
3D Gaussian Splatting (3DGS) \cite{kerbl20233dgaussiansplattingrealtime} has recently emerged as a powerful and efficient alternative scene representation for SLAM. Its explicit, point-based, differentiable nature enables high-quality, real-time rendering and facilitates optimization. Initial works demonstrated 3DGS viability for online reconstruction, primarily with RGB-D or LiDAR input \cite{matsuki2024gaussiansplattingslam, hong2024livgaussmaplidarinertialvisualfusionrealtime, ha2024rgbdgsicpslam, deng2024compact3dgaussiansplatting, hu2024cgslamefficientdensergbd}. Subsequent efforts adapted 3DGS for SLAM, including SplaTAM \cite{keetha2024splatamsplattrack} and Photo-SLAM \cite{huang2024photoslamrealtimesimultaneouslocalization}. More pertinent to our work are RGB-only 3DGS SLAM systems like Splat-SLAM \cite{sandström2024splatslamgloballyoptimizedrgbonly} and DROID-Splat \cite{homeyer2024droidsplatcombiningendtoendslam}. While achieving impressive rendering, these methods often construct the Gaussian map using proxy depth predominantly derived from the underlying visual odometry (e.g., DROID-SLAM \cite{teed2022droidslamdeepvisualslam}) or simple filtering, without explicitly modeling or leveraging confidence in different depth sources. This can lead to geometric inaccuracies, particularly when multi-view constraints are weak or monocular cues are noisy.

ConfidentSplat addresses this identified gap by introducing a dedicated confidence-aware depth fusion module within a 3DGS SLAM pipeline. Unlike prior works relying on implicit depth derivation or rudimentary proxy depth construction, our method explicitly estimates the reliability of both multi-view geometric cues and learned monocular priors, enabling a more robust and geometrically accurate scene representation by judiciously weighting their contributions.

\section{Methodology}

ConfidentSplat introduces a principled framework for robust depth estimation in 3D Gaussian Splatting (3DGS)-based Simultaneous Localization and Mapping (SLAM). This is achieved by a confidence-aware fusion of multi-view geometric constraints and learned monocular priors, enhancing the geometric fidelity and robustness of the scene representation.

\subsection{System Architecture}

\begin{figure*}[!htb]
    \centering
    \includegraphics[width=0.99\textwidth, keepaspectratio]{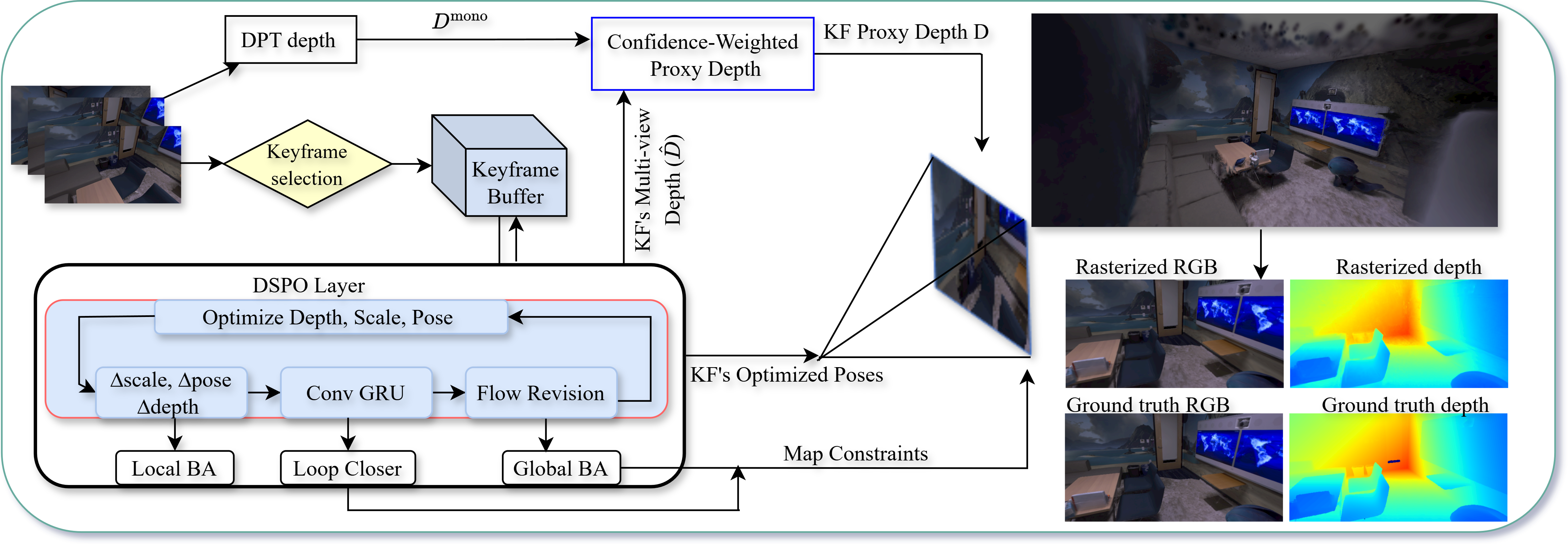} 
    \caption{The DSPO layer decomposes factor graph optimization for depth/pose refinement. Confidence fusion combines multi-view depth and monocular priors using explicit reliability weights ($w_{\text{mv}},w_{\text{mono}}$ ). Backend optimizations (loop closure, global BA) warp the 3DGS map via pose updates, ensuring global consistency.}
    \label{fig:framework}
\end{figure*}

The system employs a parallelized architecture with concurrent tracking and mapping threads (Fig. \ref{fig:framework}). The tracking thread estimates camera pose \( \boldsymbol{\omega} \in SE{3} \) and computes dense inter-keyframe correspondences and inverse depth estimates \(d \in R^+\). Concurrently, the mapping thread initializes and refines a deformable 3D Gaussian scene representation \cite{kerbl20233dgaussiansplattingrealtime}.

\subsubsection{Tracking Module}
The tracking module incrementally estimates the camera trajectory. Initial frame-to-frame motion and dense correspondences \( (\mathbf{p}_i, \tilde{\mathbf{p}}_{ij}) \), derived from a pre-trained RAFT network \cite{teed2020raftrecurrentallpairsfield} with associated confidence-encoding covariance matrices \( \mathbf{\Sigma}_{ij}(\mathbf{p}_i) \), inform a Disparity, Scale, and Pose Optimization (DSPO) layer adapted from \cite{zhang2024glorieslamgloballyoptimizedrgbonly}. Operating over a sliding window of keyframes, the DSPO layer jointly refines camera poses \( \boldsymbol{\omega}_k \), a subset of per-pixel inverse depths identified as high-error \( d^h_k \), and scale (\( \theta_k \)) and shift (\( \gamma_k \)) parameters aligning multi-view geometry with monocular priors.

This joint optimization occurs over a factor graph \( G = (V, E) \), where vertices \( V \) represent keyframe states and edges \( E \) encode geometric constraints. Keyframes are incorporated into \( G \) if the mean optical flow magnitude relative to the last keyframe surpasses a threshold \( \tau \). Optimization employs the Gauss-Newton algorithm to minimize two interleaved objectives:

First, a geometric objective enforces multi-view consistency by penalizing discrepancies between optical flow predictions \( \tilde{\mathbf{p}}_{ij} \) and projections derived from the current geometry, weighted by flow confidence:
\begin{align}
\label{eq:geom_objective_final_v2}
\mathcal{L}_{\text{geom}} = & \sum_{(i,j) \in E} \sum_{\mathbf{p}_i \in \Omega_i}
\bigg\|
\tilde{\mathbf{p}}_{ij}(\mathbf{p}_i) \nonumber \\
& - \Pi\left(
\boldsymbol{\omega}_j^{-1} \boldsymbol{\omega}_i \left(
\frac{1}{d_i(\mathbf{p}_i)} \mathbf{K}^{-1} [\mathbf{p}_i; 1]
\right)
\right)
\bigg\|^2_{\mathbf{\Sigma}_{ij}(\mathbf{p}_i)}.
\end{align}
Here, \( \Pi: R^3 \to R^2 \) is the camera projection function (intrinsics \( \mathbf{K} \)), and \( \| \cdot \|^2_{\mathbf{\Sigma}} \) denotes the squared Mahalanobis distance.

Second, to enhance robustness, dense monocular depth priors \( D^{\text{mono}} \) (from DPT \cite{eftekhar2021omnidatascalablepipelinemaking}) are integrated. This facilitates joint optimization of per-keyframe scale \( \theta_i \) and shift \( \gamma_i \) parameters, alongside the high-error inverse depths \( d^h_i \) (\( \mathbf{p} \in \text{HighErr}_i \)):
\begin{align}
\label{eq:prior_objective_final_v2}
\min_{\{d^h_k\}, \{\theta_k\}, \{\gamma_k\}} \Bigg[ &
    \mathcal{L}_{\text{geom}}(d^h) \nonumber \\
+ &\alpha_1 \sum_{i\in V} \sum_{\mathbf{p} \in \text{HighErr}_i} \| d_i^h(\mathbf{p}) - d^{\text{prior}}_i(\mathbf{p}) \|^2 \\
+ &\alpha_2 \sum_{i\in V} \sum_{\mathbf{p} \in \text{LowErr}_i} \| d_i^l(\mathbf{p}) - d^{\text{prior}}_i(\mathbf{p}) \|^2 \Bigg] \nonumber,
\end{align}
where \( d^{\text{prior}}_i(\mathbf{p}) = \theta_i (1/D^{\text{mono}}_i(\mathbf{p})) + \gamma_i \) is the scale-aligned monocular prior inverse depth.

Inverse depth estimates are classified as high-error ($d^h$) or low-error ($d^l$) by assessing multi-view geometric consistency. A consistency score \(n_i(\mathbf{p})\), reflecting alignment with reconstructions from neighboring keyframes, determines this classification: pixels with \( n_i(\mathbf{p}) > \tau_{\text{consistency}} \) are deemed low-error (\(d^l_i\)). The monocular prior in Eq.~\eqref{eq:prior_objective_final_v2} predominantly influences \(d^h\) estimates (via \(\alpha_1\)), while \(d^l\) estimates primarily guide the fitting of \( \theta_i, \gamma_i \) (via \(\alpha_2 > \alpha_1\)). Initial \( (\theta_i, \gamma_i) \) are obtained by least-squares fitting of \(d^{\text{prior}}_i\) to \(d^l_i\):
\begin{equation}
\label{eq:init_theta_gamma_final_v2}
\{\theta_i, \gamma_i\}_{\text{init}} = \mathop{\arg \min}_{\theta, \gamma}
\sum_{\mathbf{p} \in \text{LowErr}_i}
\left( d^{\text{prior}}_i(\mathbf{p}) - d_i^l(\mathbf{p}) \right)^2
\end{equation}
The tracking module outputs optimized poses \( \{\boldsymbol{\omega}_i\} \) and refined multi-view inverse depth maps \( \{\hat{d}_i\} \).

\subsubsection{Pose Graph Optimization and Consistency}
Long-term trajectory accuracy and global map coherence are ensured by backend optimization on \( G \). Loop closures, identified by geometric co-visibility and temporal separation criteria, introduce spatial constraints into \( G \), often enforced immediately through local bundle adjustment. Periodically, global Bundle Adjustment (BA) optimizes the entire factor graph \( G \), incorporating all odometry and loop closure constraints \cite{zhang2023goslamglobaloptimizationconsistent, zhang2024glorieslamgloballyoptimizedrgbonly}. For enhanced numerical stability, inverse depths \(d\) and pose translations \(\mathbf{t}\) are normalized relative to the mean inverse depth \( \overline{d} \) (i.e., \( d_{\text{norm}} = d / \overline{d} \), \( \mathbf{t}_{\text{norm}} = \overline{d} \cdot \mathbf{t} \)) before each global BA.

\subsection{Deformable 3D Gaussian Scene Representation}
We employ 3DGS \cite{kerbl20233dgaussiansplattingrealtime} for dense scene representation. Each Gaussian \( \mathcal{G}_j \) is parameterized by its mean \( \bm{\mu}_j \), covariance \( \mathbf{\Sigma}_j = \mathbf{R}_j \mathbf{S}_j \mathbf{S}_j^T \mathbf{R}_j^T \) (rotation \( \mathbf{R}_j \), scale \( \mathbf{S}_j = \text{diag}(\mathbf{s}_j) \)), opacity \( \sigma_j \), and Spherical Harmonics (SH) coefficients. A key advantage of 3DGS is its amenability to efficient deformation \cite{sandström2024splatslamgloballyoptimizedrgbonly, kong2024dgsslamgaussiansplattingslam}; pose corrections from backend optimizations are propagated to Gaussian parameters (primarily \( \bm{\mu}_j \)) to warp the map, maintaining geometric consistency with the optimized trajectory.

\subsection{Confidence-Weighted Proxy Depth Fusion}
Reliable depth supervision for 3DGS map optimization is generated by a confidence-weighted fusion mechanism. This produces a high-fidelity proxy depth map \( D_i \) for each keyframe \(i\) by adaptively blending multi-view estimates \( \hat{D}_i \) and scale-aligned monocular priors \( D^{\text{mono}}_{\text{scaled}, i} \):
\begin{equation}
    \label{eq:fusion_final}
    D_i(\mathbf{p}) = w_{\text{mv}}(\mathbf{p}) \cdot \hat{D}_i(\mathbf{p}) + w_{\text{mono}}(\mathbf{p}) \cdot D^{\text{mono}}_{\text{scaled}, i}(\mathbf{p}) \enspace,
\end{equation}
where \( D^{\text{mono}}_{\text{scaled}, i} \) is the prior aligned using \( (\theta_i, \gamma_i) \) from DSPO, and weights satisfy \( w_{\text{mv}} + w_{\text{mono}} = 1 \).

The multi-view confidence \( w_{\text{mv}}(\mathbf{p}) \) is derived from the geometric consistency score \( n_i(\mathbf{p}) \) (used for HighErr/LowErr classification), which quantifies the number of neighboring keyframes \( k \) whose reconstruction \( \mathbf{X}_k(\mathbf{p}'_k) \) aligns with \( \mathbf{X}_i(\mathbf{p}) \) from keyframe \( i \):
\begin{equation}
    \label{eq:consistency_count_fusion_later}
    n_i(\mathbf{p}) = \sum_{k \in \text{Neighbors}(i)} \mathbb{I}\left( \| \mathbf{X}_i(\mathbf{p}) - \mathbf{X}_k(\mathbf{p}'_k) \|_2 < \eta \cdot \text{mean}(\hat{D}_i) \right) \enspace.
\end{equation}
Normalizing this score yields the confidence weight:
\begin{equation}
    \label{eq:wmv_norm_final}
    w_{\text{mv}}(\mathbf{p}) = \frac{n_i(\mathbf{p})}{N_{\text{key}}} \enspace,
\end{equation}
where \( N_{\text{key}} \) is the number of keyframes considered. Monocular confidence is complementary:
\begin{equation}
    \label{eq:wmono_complement_final}
    w_{\text{mono}}(\mathbf{p}) = 1 - w_{\text{mv}}(\mathbf{p}) \enspace.
\end{equation}
This ensures increased prior influence when multi-view evidence is weak. This adaptive fusion strategy generates robust proxy depth \( D \) for superior geometric guidance.

\subsection{3D Gaussian Map Initialization and Optimization}
The mapping thread utilizes refined poses \( \{\boldsymbol{\omega}_i\} \) and confidence-weighted proxy depth maps \( \{D_i\} \) to construct and optimize the 3DGS representation \( \mathcal{G} \).

\subsubsection{Map Initialization}
New Gaussian primitives are instantiated for new keyframes. For selected pixels \( \mathbf{p} \), 3D positions \( \mathbf{X}_i(\mathbf{p}) \), computed by unprojecting \( \mathbf{p} \) using \( \boldsymbol{\omega}_i \) and the fused proxy depth \( D_i(\mathbf{p}) \), serve as initial means \( \bm{\mu} \). Other parameters are set heuristically or from image data \cite{matsuki2024gaussiansplattingslam, sandström2024splatslamgloballyoptimizedrgbonly}.

\subsubsection{Map Optimization and Deformation}
Gaussian parameters \( \{\bm{\mu}_j, \mathbf{\Sigma}_j, \sigma_j, \mathbf{c}_j\} \) are continuously optimized by minimizing a composite loss over supervising keyframes (KFs):
\begin{align}
\label{eq:map_loss_final}
\mathcal{L} = & \sum_{k \in \text{KFs}} \sum_{\mathbf{p} \in \Omega_k} \left[ (1-\lambda_{\text{SSIM}})\mathcal{L}_{1} + \lambda_{\text{SSIM}}\mathcal{L}_{\text{SSIM}} \right](C'(\mathbf{p}), C_k(\mathbf{p})) \notag \\
& + \lambda_{\text{depth}} \sum_{k \in \text{KFs}} \sum_{\mathbf{p} \in \Omega_k} \| D^{\text{rendered}}(\mathbf{p}) - D^{\text{proxy}}(\mathbf{p}) \|_1 \notag \\
& + \lambda_{\text{reg}} \sum_j \mathcal{R}(\mathbf{\Sigma}_j) \enspace.
\end{align}
This loss includes: (1) a photometric term (L1 and SSIM) comparing rendered color \( C' \) with observed color \( C_k \); (2) abyometric term, weighted by \( \lambda_{\text{depth}} \), enforcing consistency between rendered depth \( D^{\text{rendered}} \) and our confidence-weighted proxy depth \(D^{\text{proxy}} \); and (3) a regularization term \( \mathcal{R}(\mathbf{\Sigma}_j) \). Optimization uses stochastic gradient descent. The map undergoes non-rigid deformation based on pose updates from the backend to maintain global consistency \cite{sandström2024splatslamgloballyoptimizedrgbonly}.

\begin{figure*}[t]
    \centering \hspace{-3em}
    {
        \setlength{\tabcolsep}{0.5pt} 
        \renewcommand{\arraystretch}{1.5} 
        \newcommand{\sz}{0.28} 
        \newcommand{\hgt}{3.2cm} 
        \begin{tabular}{@{}ccccc@{}}
            \tiny\bfseries
            \raisebox{1cm}{\rotatebox{90}{\makecell{TUM RGB-D\\\texttt{fr1/desk}}}} &
            \includegraphics[width=\sz\linewidth, height=\hgt, keepaspectratio]{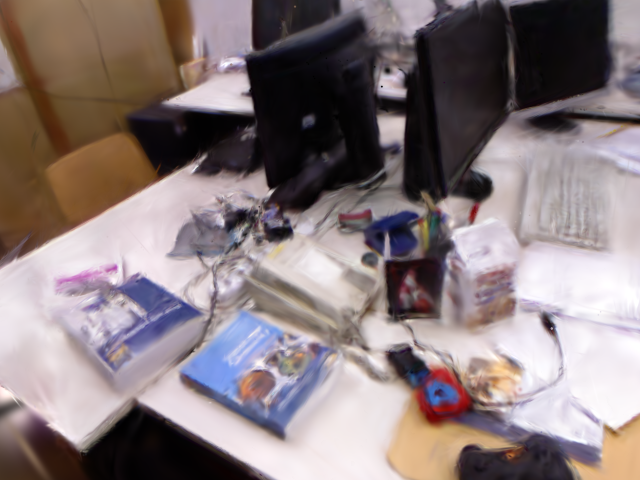} &
            \includegraphics[width=\sz\linewidth, height=\hgt, keepaspectratio, trim=15 12 15 12, clip]{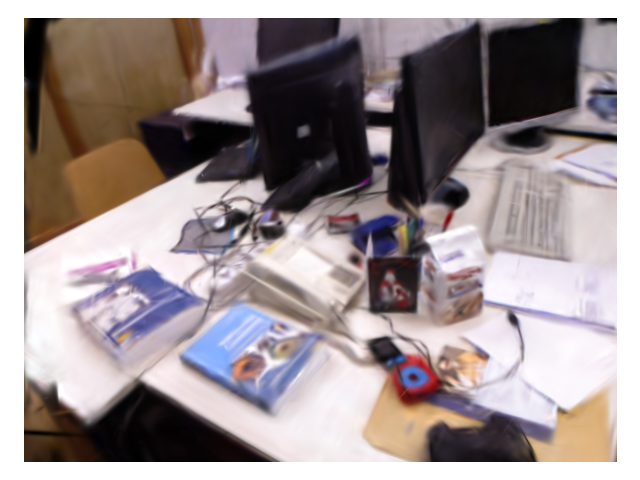} &
            \includegraphics[width=\sz\linewidth, height=\hgt, keepaspectratio, trim=5 10 5 10, clip]{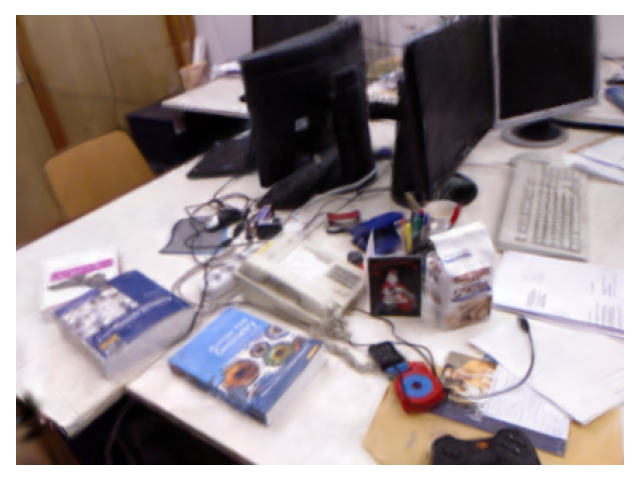} &
            \includegraphics[width=\sz\linewidth, height=\hgt, keepaspectratio]{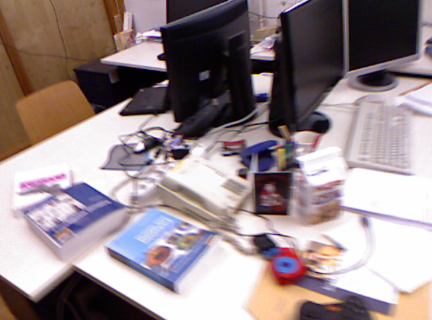} \\
            &
            \includegraphics[width=\sz\linewidth, height=\hgt, keepaspectratio]{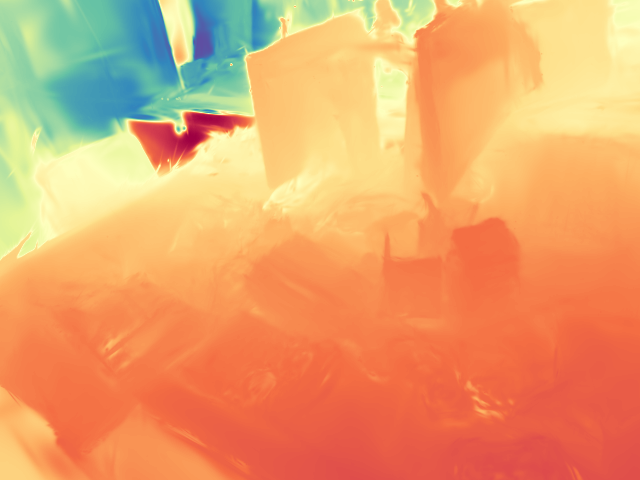} &
            \includegraphics[width=\sz\linewidth, height=\hgt, keepaspectratio, trim=15 12 15 12, clip]{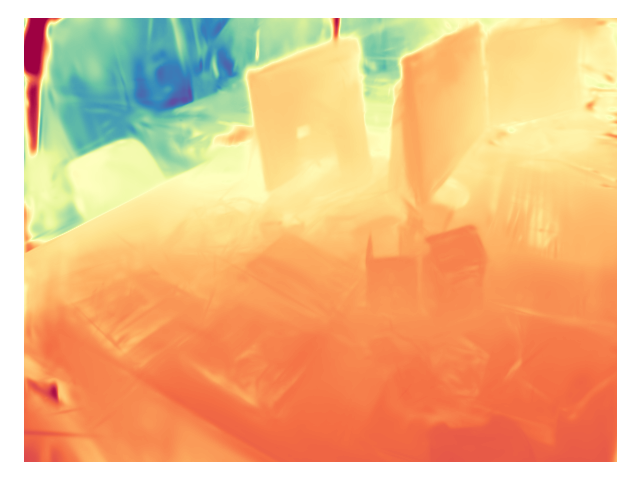} &
            \includegraphics[width=\sz\linewidth, height=\hgt, keepaspectratio, trim=5 10 5 10, clip]{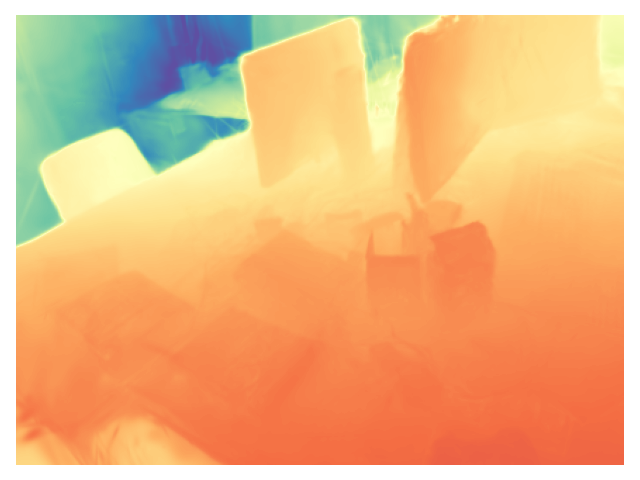} &
            \includegraphics[width=\sz\linewidth, height=\hgt, keepaspectratio]{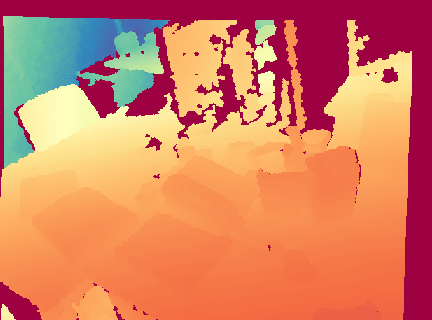} \\
            & \scriptsize Photo-SLAM~\cite{huang2024photoslamrealtimesimultaneouslocalization} & \scriptsize MonoGS~\cite{Matsuki:Murai:etal:CVPR2024} & \scriptsize \textbf{ConfidentSplat (Ours)} & \scriptsize Ground Truth \\[-4pt]
        \end{tabular}
    }
    \caption{Qualitative comparison on TUM-RGBD (\texttt{fr1/desk}). ConfidentSplat yields improved novel view synthesis (top) and depth estimation (bottom) compared to representative baselines. Enhancements stem from our confidence-aware fusion mechanism operating on dense tracking outputs.}
    \label{fig:render_tum}
\end{figure*}

\section{Experimental Evaluation}

We conducted a rigorous empirical validation of ConfidentSplat, performing comprehensive comparisons against state-of-the-art methods on standard benchmarks and diverse custom data. To isolate the specific contributions of our proposed fusion mechanism, we conduct detailed analysis against the Splat-SLAM baseline \cite{sandström2024splatslamgloballyoptimizedrgbonly} on custom data.

\subsection{Experimental Setup}

\subsubsection{Datasets}
Evaluation employed four datasets:
\textbf{(1) TUM RGB-D}~\cite{6385773}: Standard indoor sequences (\texttt{fr1/desk}, \texttt{fr2/xyz}, \texttt{fr3/office}, \texttt{fr1/desk2}, \texttt{fr1/room}).
\textbf{(2) ScanNet v2}~\cite{dai2017scannetrichlyannotated3dreconstructions}: Large-scale indoor reconstructions (\texttt{scene0000}, \texttt{0059}, \texttt{0106}, \texttt{0181}, \texttt{0207}, `\_00' versions).
\textbf{(3) Replica}~\cite{straub2019replicadatasetdigitalreplica}: High-quality synthetic indoor scenes (8 scenes evaluated).
\textbf{(4) Custom}: Challenging indoor sequences captured via mobile phone, pre-processed using COLMAP.

\subsubsection{Evaluation Metrics}
Performance was quantified across three domains:
(1) \textbf{Rendering Quality}: PSNR$\uparrow$, SSIM$\uparrow$ \cite{zhang2018unreasonableeffectivenessdeepfeatures}, and LPIPS$\downarrow$ \cite{zhang2018unreasonableeffectivenessdeepfeatures}.
(2) \textbf{Reconstruction Accuracy}: Depth L1 Error$\downarrow$, geometric Accuracy$\downarrow$ (Chamfer-L1), Completion$\downarrow$ (Chamfer-L1), and Completion Ratio$\uparrow$, following \cite{zhu2023nicerslamneuralimplicitscene}.
(3) \textbf{Tracking Accuracy}: ATE RMSE$\downarrow$ \cite{6385773}. System efficiency is noted via Map Size (MB)$\downarrow$ and Runtime (FPS)$\uparrow$.

\subsection{Implementation Details}

Our system, implemented in Python with PyTorch , extends the Splat-SLAM \cite{sandström2024splatslamgloballyoptimizedrgbonly} architecture. It integrates a DROID-SLAM-inspired \cite{teed2022droidslamdeepvisualslam} tracker using RAFT \cite{teed2020raftrecurrentallpairsfield} flow and Omnidata \cite{eftekhar2021omnidatascalablepipelinemaking} depth priors, with a 3DGS \cite{kerbl20233dgaussiansplattingrealtime} mapper enhanced by our confidence-weighted fusion. Empirically determined hyperparameters are listed in Table \ref{table_hyperparams}. Experiments were performed on an NVIDIA A100 GPU.

\begin{table}
\caption{Core Hyperparameters.}
\begin{center}
\begin{tabular}{|c|c|}
\hline
\textbf{Hyperparameter} & \textbf{Value} \\
\hline
Consistency Thresh. (\(\eta\)) & 0.01 \\
\hline
Opt. Keyframes (\(N_{\text{key}}\)) & 30 \\
\hline
Depth Loss Weight (\(\lambda_{\text{depth}}\)) & 0.2 \\
\hline
Regularization (\(\lambda_{\text{reg}}\)) & 10.0 \\
\hline
\end{tabular}
\label{table_hyperparams}
\end{center}
\end{table}

\begin{table}
\caption{Tracking accuracy (ATE RMSE [cm] $\downarrow$) on TUM-RGBD benchmarks \cite{6385773}. Lower is better. Baseline results from respective publications. \textbf{Best among listed methods} highlighted.}
\begin{center}
\begin{tabular}{|l|c|c|c|c|c|c|}
\hline
\textbf{Method} & \texttt{dsk} & \texttt{xyz} & \texttt{off} & \texttt{dsk2} & \texttt{rm} & \textbf{Avg.} \\
\hline
DPV-SLAM \cite{lipson2024deeppatchvisualslam} & 1.8 & 1.0 & - & 2.9 & 9.6 & - \\
\hline
GlORIE \cite{zhang2024glorieslamgloballyoptimizedrgbonly} & 1.6 & \textbf{0.2} & 1.4 & \textbf{2.8} & \textbf{4.2} & \textbf{2.1} \\
\hline
GO-SLAM \cite{zhang2023goslamglobaloptimizationconsistent} & 1.6 & 0.6 & 1.5 & \textbf{2.8} & 5.2 & 2.3 \\
\hline
MonoGS \cite{Matsuki:Murai:etal:CVPR2024} & 4.2 & 4.8 & 4.4 & - & - & - \\
\hline
MoD-SLAM \cite{zhou2024modslammonoculardensemapping} & \textbf{1.5} & 0.7 & \textbf{1.1} & - & - & - \\
\hline
Photo-SLAM \cite{huang2024photoslamrealtimesimultaneouslocalization} & \textbf{1.5} & 1.0 & 1.3 & - & - & - \\
\hline
\textbf{Ours} & 1.6 & \textbf{0.2} & 1.6 & 8.3 & 5.8 & 3.5 \\
\hline
\end{tabular}
\label{tab:sota_tracker}
\end{center}
\end{table}

\begin{table}
\caption{Tracking accuracy comparison (Error [m] $\downarrow$) on custom data vs. baseline \cite{sandström2024splatslamgloballyoptimizedrgbonly}. \textbf{Ours} shows improvement.}
\begin{center}
\begin{tabular}{|l|c|c|}
\hline
\textbf{Metric} & Splat-SLAM~\cite{sandström2024splatslamgloballyoptimizedrgbonly} & \textbf{Ours} \\
\hline
RMSE & 3.8686 & \textbf{3.7059} \\
\hline
Mean & 3.7478 & \textbf{3.4500} \\
\hline
Median & 3.6757 & \textbf{3.2892} \\
\hline
\end{tabular}
\label{tab:full_trajectory_comparison}
\end{center}
\end{table}

\subsection{Results and Analysis}

\paragraph{Tracking Performance:}

Comparative tracking evaluation on TUM-RGBD (Table \ref{tab:sota_tracker}) shows ConfidentSplat achieves state-of-the-art accuracy, matching GlORIE on the challenging \texttt{fr2/xyz} sequence, and remains competitive on \texttt{fr1/desk}. However, performance degrades on longer sequences relative to methods like GlORIE \cite{zhang2024glorieslamgloballyoptimizedrgbonly} and GO-SLAM \cite{zhang2023goslamglobaloptimizationconsistent}. This suggests sensitivity to accumulated drift, potentially exacerbated by our reliance on a fixed, pre-trained tracking frontend. Critically, when evaluated against its direct baseline (Splat-SLAM\cite{sandström2024splatslamgloballyoptimizedrgbonly}) on our custom dataset (Table \ref{tab:full_trajectory_comparison}), ConfidentSplat demonstrates markedly superior accuracy across all error metrics (RMSE, Mean, Median). This indicates that our confidence-aware fusion mechanism significantly enhances robustness, particularly in diverse conditions potentially underrepresented in standard benchmarks where the pre-trained tracker might be less optimal. While the fusion clearly improves results relative to the baseline in these scenarios, the absolute tracking accuracy on custom data may still be influenced by the tracker operating outside its original training distribution.

\begin{figure*}[t]
    \centering \vspace{-1mm}
    {
        \setlength{\tabcolsep}{1pt}
        \renewcommand{\arraystretch}{1.0}
        \newcommand{\sz}{0.49}
        \begin{tabular}{@{}cc@{}}
            \includegraphics[width=\sz\linewidth, keepaspectratio]{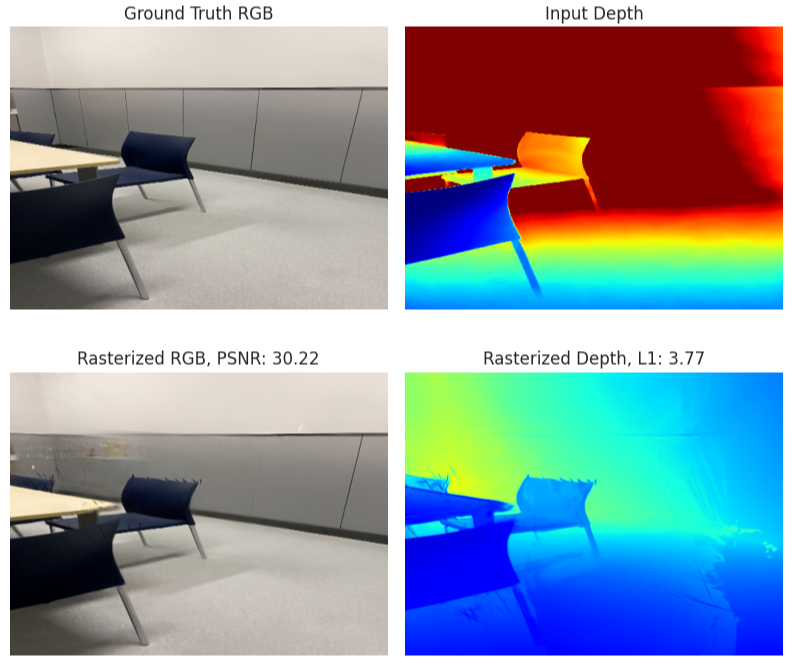} &
            \includegraphics[width=\sz\linewidth, keepaspectratio]{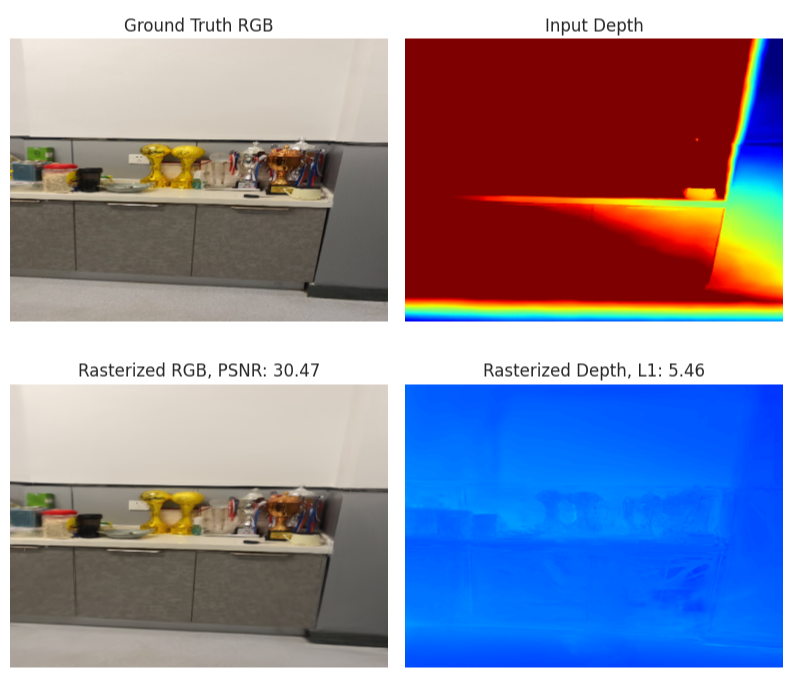}
        \end{tabular}
    }
    \caption{High-fidelity rendering on challenging custom data (mobile phone capture). ConfidentSplat effectively reconstructs complex geometry and appearance, validating performance in diverse, uncontrolled settings.}
    \label{custom_data1} \vspace{-3mm}
\end{figure*}

\begin{table}[t] 
\caption{Novel view synthesis quality on ScanNet v2 \cite{dai2017scannetrichlyannotated3dreconstructions}. \textbf{Best among listed methods} highlighted.}
\begin{center}
\setlength{\tabcolsep}{3pt} 
\begin{tabular}{|l|c|c|c|c|c|c|c|c|}
\hline
\textbf{Method} & \textbf{Metric} & \texttt{0000} & \texttt{0059} & \texttt{0106} & \texttt{0181} & \texttt{0207} & \textbf{Avg.} \\
\hline
\multirow{3}{*}{\makecell{GO-\\SLAM~\cite{zhang2023goslamglobaloptimizationconsistent}} }
  & PSNR $\uparrow$ & 15.74 & 13.15 & 14.58 & 15.72 & 15.37 & 14.91 \\
\cline{2-8} 
  & SSIM $\uparrow$ & 0.42 & 0.32 & 0.46 & 0.53 & 0.39 & 0.42 \\
\cline{2-8}
  & LPIPS $\downarrow$ & 0.61 & 0.60 & 0.59 & 0.62 & 0.60 & 0.60 \\
\hline
\multirow{3}{*}{\makecell{GlORIE-\\SLAM~\cite{zhang2024glorieslamgloballyoptimizedrgbonly}} }
  & PSNR $\uparrow$ & 23.42 & 20.66 & 20.41 & 25.23 & 23.68 & 22.68 \\
\cline{2-8}
  & SSIM $\uparrow$ & \textbf{0.87} & \textbf{0.87} & 0.83 & \textbf{0.84} & 0.76 & \textbf{0.83} \\
\cline{2-8}
  & LPIPS $\downarrow$ & 0.26 & 0.31 & 0.31 & 0.21 & 0.29 & 0.27 \\
\hline
\multirow{3}{*}{MonoGS~\cite{Matsuki:Murai:etal:CVPR2024}} 
  & PSNR $\uparrow$ & 16.91 & 19.15 & 18.57 & 19.51 & 18.37 & 18.50 \\
\cline{2-8}
  & SSIM $\uparrow$ & 0.62 & 0.69 & 0.74 & 0.75 & 0.70 & 0.70 \\
\cline{2-8}
  & LPIPS $\downarrow$ & 0.70 & 0.51 & 0.55 & 0.63 & 0.58 & 0.59 \\
\hline
\multirow{3}{*}{\textbf{Ours}} 
  & PSNR $\uparrow$ & \textbf{26.68} & \textbf{24.96} & \textbf{26.64} & \textbf{29.11} & \textbf{31.53} & \textbf{27.78} \\
\cline{2-8}
  & SSIM $\uparrow$ & 0.56 & 0.77 & \textbf{0.86} & 0.82 & \textbf{0.81} & 0.76 \\
\cline{2-8}
  & LPIPS $\downarrow$ & 0.48 & 0.55 & 0.48 & 0.25 & \textbf{0.29} & 0.41 \\
\hline
\end{tabular}
\label{table_scanNet}
\end{center}
\end{table}

\begin{table}
\caption{Reconstruction accuracy [cm] on Replica \cite{straub2019replicadatasetdigitalreplica} (RGB methods avg. 8 scenes). Lower is better ($\downarrow$), except Comp. Rat. ($\uparrow$)}
\begin{center}
\setlength{\tabcolsep}{3pt} 
\begin{tabular}{|l|c|c|c|c|c|c|}
\hline
\textbf{Metric} & \rotatebox{60}{DIM-S \cite{li2023densergbslamneural}} & \rotatebox{60}{GO-S \cite{zhang2023goslamglobaloptimizationconsistent}} & \rotatebox{60}{NICER-S \cite{zhu2023nicerslamneuralimplicitscene}} & \rotatebox{60}{HI-S \cite{zhang2023hislammonocularrealtimedense}} & \rotatebox{60}{MonoGS \cite{Matsuki:Murai:etal:CVPR2024}} & \rotatebox{60}{\textbf{Ours}} \\
\hline
Depth L1 $\downarrow$ & - & - & - & - & 27.24 & \textbf{2.37} \\
\hline
Acc. $\downarrow$ & 4.03 & 3.81 & 3.65 & 3.62 & 30.61 & 3.18 \\
\hline
Compl. $\downarrow$ & 4.20 & 4.79 & 4.16 & 4.59 & 12.19 & 4.12 \\
\hline
Comp. Rat. $\uparrow$ & 79.6 & 78.0 & 79.4 & 80.6 & 40.5 & 78.6 \\
\hline
\end{tabular}
\label{table_Replica}
\end{center}
\end{table}

\begin{table}
\caption{Rendering quality and depth error on custom data vs. baseline \cite{sandström2024splatslamgloballyoptimizedrgbonly}. \textbf{Ours} shows improvement.}
\begin{center}
\setlength{\tabcolsep}{4pt} 
\begin{tabular}{|l|c|c|}
\hline
\textbf{Metric} & Splat-SLAM~\cite{sandström2024splatslamgloballyoptimizedrgbonly} & \textbf{Ours} \\
\hline
PSNR (dB) $\uparrow$ & 28.82 & \textbf{32.74} \\
\hline
SSIM $\uparrow$ & 0.8942 & \textbf{0.9306} \\
\hline
LPIPS $\downarrow$ & 0.1831 & \textbf{0.1252} \\
\hline
Depth L1 (m) $\downarrow$ & 5.4264 & \textbf{4.9011} \\
\hline
\end{tabular}
\label{tab:image_quality_comparison}
\end{center}
\end{table}

\paragraph{Rendering and Reconstruction Performance:}
ConfidentSplat demonstrates strong rendering capabilities. On ScanNet (Table \ref{table_scanNet}), it achieves the best average PSNR among the compared methods and leads in specific scenes for PSNR and SSIM, although GlORIE-SLAM shows better average SSIM. On the synthetic Replica dataset (Table \ref{table_Replica}), our method sets a new state-of-the-art for Render Depth L1 accuracy among evaluated monocular techniques, while maintaining competitive geometric accuracy and completion against implicit representation methods like GlORIE-SLAM \cite{zhang2024glorieslamgloballyoptimizedrgbonly}. The most compelling evidence for the efficacy of our fusion mechanism comes from the custom dataset evaluations (Tables \ref{tab:image_quality_comparison}, \ref{tab:depth_comparison}). Here, ConfidentSplat consistently and substantially outperforms its direct baseline, Splat-SLAM, across all rendering quality metrics (PSNR, SSIM, LPIPS) and achieves significantly reduced depth estimation errors, both overall and particularly for near-field geometry ($\leq$ 4m). This robust performance uplift in challenging, real-world capture scenarios (Fig. \ref{custom_data1}) strongly validates the practical benefits of confidence-weighted fusion for enhancing reconstruction quality and rendering fidelity.

\begin{table}
\caption{Detailed depth estimation errors (m $\downarrow$) on custom data vs. baseline \cite{sandström2024splatslamgloballyoptimizedrgbonly}. \textbf{Ours} shows improvement.}
\begin{center}
\setlength{\tabcolsep}{4pt} 
\begin{tabular}{|l|c|c|}
\hline
\textbf{Metric} & Splat-SLAM~\cite{sandström2024splatslamgloballyoptimizedrgbonly} & \textbf{Ours} \\
\hline
Overall L1 & 0.7084 & \textbf{0.6629} \\
\hline
Overall L1 (Scaled) & 3.2997 & \textbf{2.9944} \\
\hline
L1 ($\leq$ 4m) & 0.3684 & \textbf{0.3431} \\
\hline
L1 ($\leq$ 4m, Scaled) & 1.3773 & \textbf{1.3073} \\
\hline
\end{tabular}
\label{tab:depth_comparison}
\end{center}
\end{table}

\section{Ablation Study}

To isolate the causal impact of our core contribution—confidence-weighted depth fusion—we performed a direct comparison against the Splat-SLAM baseline \cite{sandström2024splatslamgloballyoptimizedrgbonly}, which is similar to our system architecture without this specific fusion mechanism. The substantial performance improvements observed on the custom dataset (Tables \ref{tab:image_quality_comparison}, \ref{tab:depth_comparison})—specifically, the significant reductions in rendering artifacts (higher PSNR/SSIM, lower LPIPS) and more accurate depth estimation—are directly attributable to the proposed fusion strategy. This comparative analysis serves as an effective ablation, demonstrating the necessity and efficacy of dynamically integrating multi-view and monocular depth cues based on estimated confidence for achieving superior downstream reconstruction and rendering outcomes.

\section{Limitations and Future Work}

Despite the demonstrated advantages, several limitations inform future research. (1) Performance heterogeneity between benchmarks and custom data motivates deeper investigation into factors governing generalization, including dataset characteristics and hyperparameter robustness. (2) A rigorous computational analysis is essential to profile the overhead induced by the confidence module and explore optimizations for real-time deployment on varied hardware. (3) Validating robustness necessitates evaluation on a wider array of challenging datasets, potentially including scenes with significant dynamic content or varied environmental conditions. (4) The fixed nature of the pre-trained tracker can constrain overall system adaptability; exploring tracker fine-tuning, alternative frontends, or integrated end-to-end training strategies represents a key avenue for improving tracking consistency and optimizing system-level performance across diverse inputs.

\section{Conclusion}

We introduced ConfidentSplat, a 3DGS-based SLAM system featuring a principled, confidence-weighted mechanism for fusing multi-view geometry with monocular depth priors. Our approach achieves competitive tracking accuracy, attains state-of-the-art depth rendering quality on Replica compared to other monocular methods, and, most importantly, demonstrates substantial, consistent improvements in reconstruction fidelity and rendering quality over its baseline on challenging custom data. This work validates the significant potential of confidence-aware sensor fusion to enhance the robustness and accuracy of dense 3D reconstruction, offering a viable path towards more reliable real-time spatial perception systems effective in diverse and uncontrolled real-world environments.

\bibliographystyle{ieeetr}
\bibliography{citation}
\vspace{12pt}
\color{red}
\end{document}